\documentclass[times,twocolumn,final,authoryear]{elsarticle}

\usepackage{ycviu}
\usepackage{framed,multirow}
\usepackage{array}
\usepackage{amssymb}
\usepackage{latexsym}

\usepackage{url}
\usepackage{xcolor}
\definecolor{newcolor}{rgb}{.8,.349,.1}

\journal{Computer Vision and Image Understanding}

\begin{document}

\begin{frontmatter}

\title{Other Tokens Matter: Exploring Global and Local Features of Vision Transformers for Object Re-Identification}

\author[1]{Yingquan \snm{Wang}}
\author[2]{Pingping \snm{Zhang}\corref{cor1}}
\cortext[cor1]{Corresponding author:
  Tel.: +0-000-000-0000;
  fax: +0-000-000-0000;}
\ead{zhpp@dlut.edu.cn}
\author[1]{Dong \snm{Wang}}
\author[1]{Huchuan \snm{Lu}}

\address[1]{School of Information and
Communication Engineering, Dalian University of Technology}
\address[2]{School of Future Technology, School of Artificial Intelligence, Dalian University of
Technology}

\received{1 May 2013}
\finalform{10 May 2013}
\accepted{13 May 2013}
\availableonline{15 May 2013}
\communicated{S. Sarkar}

\begin{abstract}
Object Re-Identification (Re-ID) aims to identify and retrieve specific objects from images captured at different places and times.
Recently, object Re-ID has achieved great success with the advances of Vision Transformers (ViT).
However, the effects of the global-local relation have not been fully explored in Transformers for object Re-ID.
In this work, we first explore the influence of global and local features of ViT and then further propose a novel Global-Local Transformer (GLTrans) for high-performance object Re-ID.
We find that the features from last few layers of ViT already have a strong representational ability, and the global and local information can mutually enhance each other.
Based on this fact, we propose a Global Aggregation Encoder (GAE)  to utilize the class tokens of the last few Transformer layers and learn comprehensive global features effectively.
Meanwhile, we propose the Local Multi-layer Fusion (LMF) which leverages both the global cues from GAE and multi-layer patch tokens to explore the discriminative local representations.
Extensive experiments demonstrate that our proposed method achieves superior performance on four object Re-ID benchmarks.
\end{abstract}


\end{frontmatter}


\section{Introduction}
\begin{figure}[htbp]
    \centering
    \includegraphics[width=1.0\linewidth]{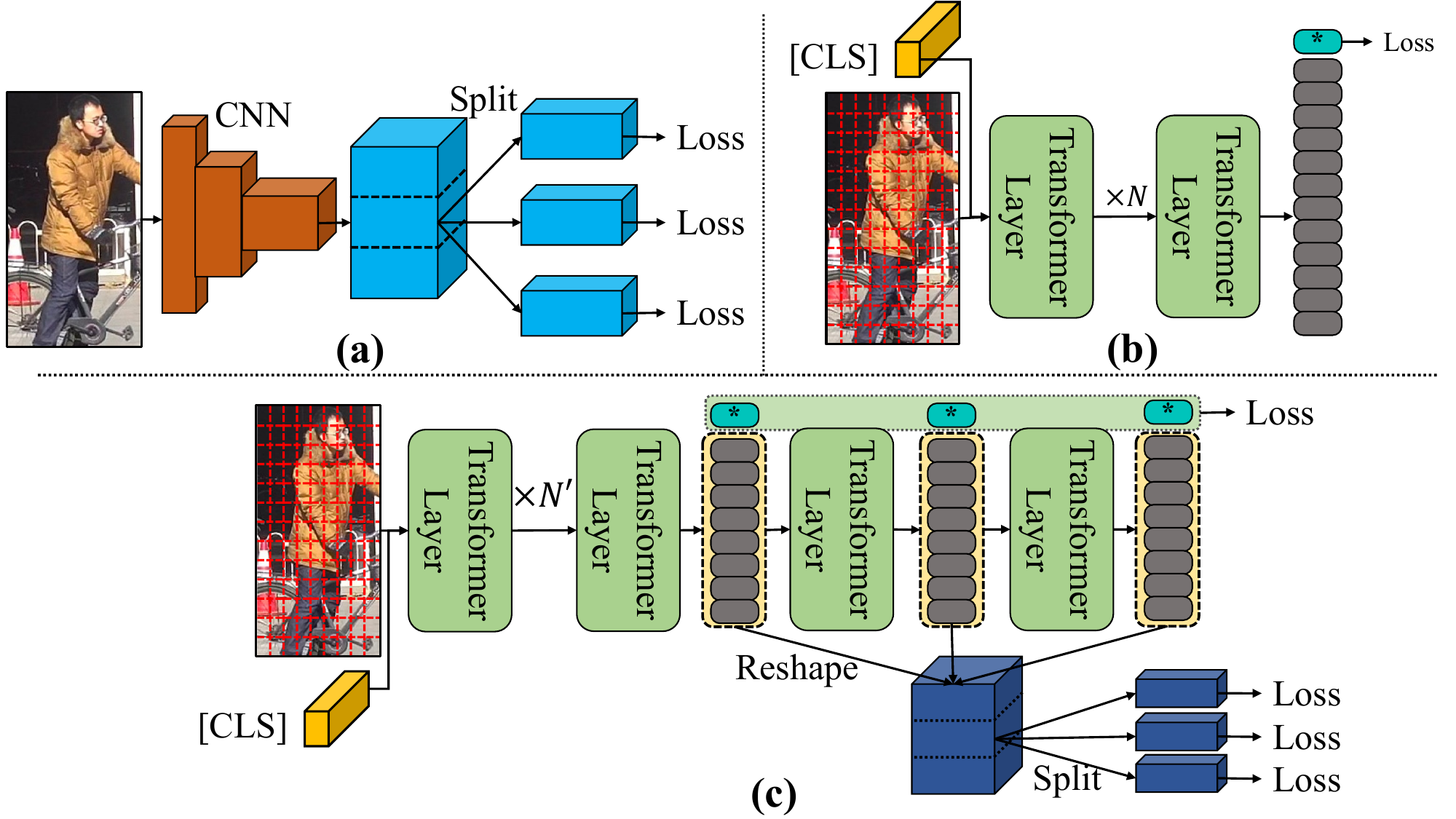}
    \caption{Different structures employed in object Re-ID. (a) Part-based CNNs for local features. (b) Pure Transformers for global features. (c) Our proposed GLTrans method considers both local and global features.}
    \label{fig:introduction}
\end{figure}

Object Re-Identification (Re-ID) aims to retrieve specific objects from images taken at different times and places.
It has drawn lots of attention due to its many real-world applications, such as safe communities, intelligent surveillance and criminal investigations~\citep{ye2021deep,zheng2016person,WENG2023103831}.
Benefiting from the local modeling ability of Convolutional Neural Networks (CNNs)~\citep{resnet}, CNN-based methods have dominated the object Re-ID~\citep{pcb,MGN,KHATUN2020102989,LIN2023103623,zhang2023multi} over the past two decades.
Among them, one of most straightforward and efficient strategies is to split the feature maps to obtain fine-grained cues~\citep{pcb,MGN}.
The main idea is shown in Fig.~\ref{fig:introduction}(a).
Although excellent performances are achieved, this kind of methods are limited by the poor global representation ability of the convolution operations~\citep{peng2021conformer}.
\begin{figure}[htbp]
  \centering
  \includegraphics[width=0.95\linewidth]{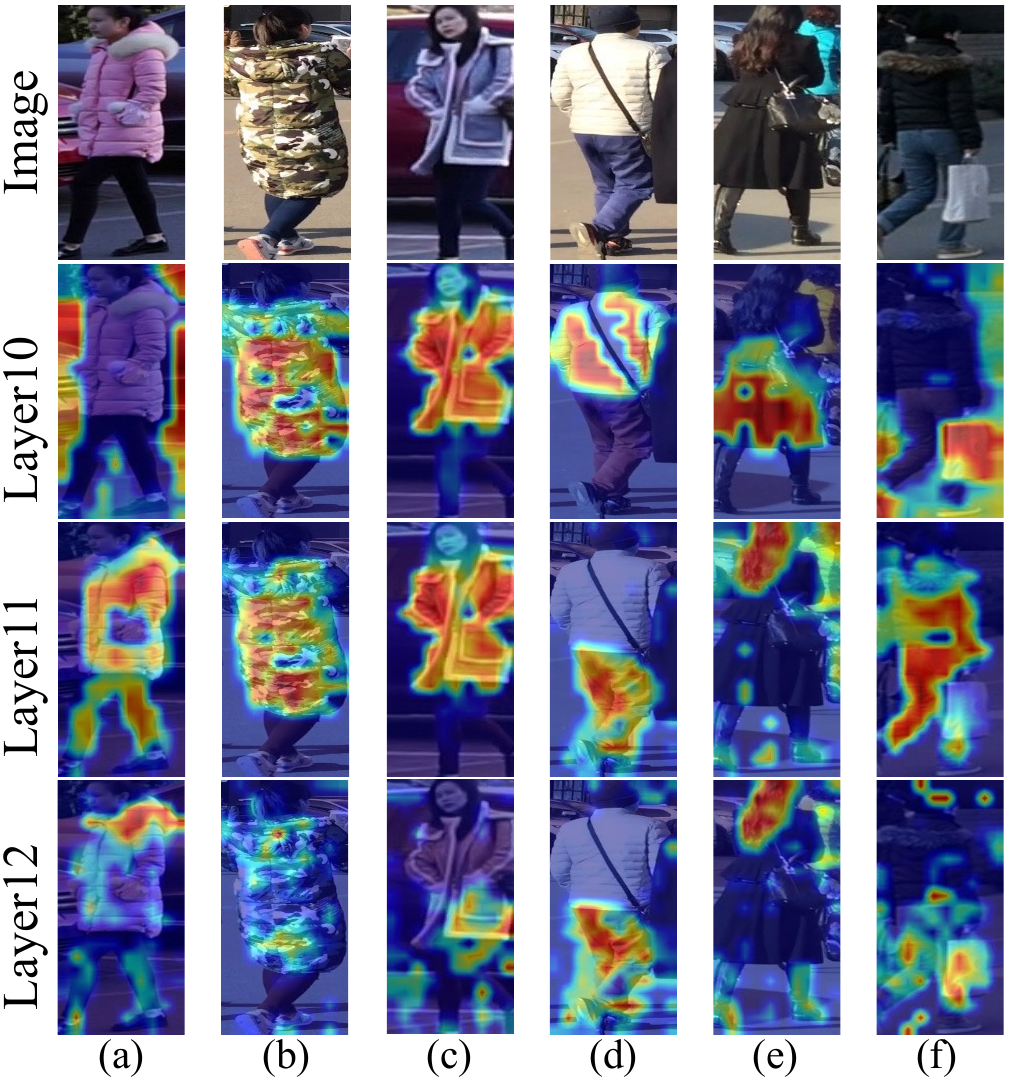}
  \caption{Heatmap visualization of ViT's different layers by Gram-Cam~\citep{gram_cam} on MSMT17. Specifically, Layer10, Layer11 and Layer12 mean the heatmap of the 10-\emph{th}, 11-\emph{th} and 12-\emph{th} layers from ViT. Deeper red colors signify higher weights.}
  \label{fig:multiple layer}
\end{figure}

Recently, Transformers as powerful structures~\citep{NIPS2017_3f5ee243} have demonstrated superior performance for many visual tasks such as image classification and object detection~\citep{NIPS2017_3f5ee243,vit}.
The key reason is that Transformers aggregate information based on multi-head self-attention and focus on long-distance dependencies~\citep{peng2021conformer}.
Inspired by this fact, He~\emph{et al.}~\citep{He_2021_ICCV} introduce the first pure Transformer-based method for object Re-ID.
Following that, many pure Transformer-based methods have been proposed~\citep{Lai_2021_ICCV,Li_2021_CVPR,Zhu_2022_CVPR, LIU2023103652}.
As illustrated in Fig.~\ref{fig:introduction} (b), the class token is employed to represent the entire image.
However, these methods usually neglect two key issues: 1) The patch tokens contain rich fine-grained cues. 2) The features from the last few layers also have strong representations.
To address these issues, some researchers divide the patch tokens into several independent regions for mining local discriminative cues.
For instance, Zhu \emph{et al.}~\citep{AAformer} employ the optimal transport algorithm~\citep{cuturi2013sinkhorn} to discover local tokens with shared semantics and subsequently extract fine-grained information.
Wang \emph{et al.}~\citep{wang2022pose} and Zang \emph{et al.}~\citep{pit} reorganize patch tokens into feature maps and split these feature maps horizontally to extract local-wise representations.
However, different from the feature map of CNNs, each patch token contains diverse global-view information, which needs to be further selected and refined.
In addition, simple partitions may miss the structure information.

On the other hand, Fig.~\ref{fig:multiple layer} visualizes the last three layers of ViT fine-tuned on the MSMT17 dataset~\citep{MSMT17}.
It is apparent that each layer emphasizes different semantics, and the patch tokens show strong representations due to the excellent global modeling of Transformers.
For example, as depicted in Fig.~\ref{fig:multiple layer} (d), the patch tokens of Layer-10 focus on the upper white coat, while that of Layer-11 and Layer-12 focus on the blue pants and shoes.
Similar observations can be found in Fig.~\ref{fig:multiple layer} (b) and Fig.~\ref{fig:multiple layer} (c).
In addition, we find that the most effective representations of patch tokens are not consistently found in the final layer.
For instance, as shown in Fig.~\ref{fig:multiple layer} (b), Fig.~\ref{fig:multiple layer} (c) and Fig.~\ref{fig:multiple layer} (f), the last layer is inferior to the shallower layers.
Thus, using solely the features from the last layer may be a suboptimal choice.
Furthermore, obtaining comprehensive fine-grained representations requires a full consideration of both the patch tokens and multi-stage features.

To this end, we propose a novel framework named Global-Local Transformer (GLTrans) to obtain a more robust and compact representation for object Re-ID.
The main structure is shown in Fig.~\ref{fig:introduction} (c), which is very different from previous works.
More specifically, we first obtain the multi-layer patch tokens and class tokens from ViT~\citep{vit}.
Then, they are passed through a Local Multi-layer Fusion (LMF) and a Global Aggregation Encoder (GAE) for generating more discriminative local and global features, respectively.
In GAE, the multi-layer class tokens are passed through a Fully-Connected (FC) layer for a comprehensive global feature.
On the other hand, the multi-layer patch tokens are fed into the LMF to mine fine-grained information.
Specifically, the LMF contains three parts: Patch Token Fusion (PTF), Global-guided Multi-head Attention (GMA) and Part-aware Transformer Layer (PTL).
Firstly, the PTF re-weights and fuses the multi-layer patch tokens.
Then the GMA further enhances patch tokens guided by global features.
Finally, the PTL generates discriminative local features from multiple regions of enhanced patch tokens.
Extensive experiments on four large-scale object Re-ID benchmarks demonstrate that our method shows better results than most state-of-the-art methods.

The main contributions are summarized as follows:
\begin{itemize}
  \item We propose a novel learning framework (\emph{i.e.}, GLTrans) to take local and global advantages of vision Transformers for robust object Re-ID.
  \item We propose the LMF to fuse multi-layer patch tokens for discriminative local representations. Additionally, we also present the GAE to aggregate multi-layer class tokens for comprehensive global representations.
  \item Extensive experiments demonstrate that our framework can effectively extract comprehensive feature representations. It achieves outstanding performances on four large-scale object Re-ID benchmarks.
\end{itemize}
\section{Related Work}
\subsection{CNNs for Object Re-Identification}
Recently, object Re-ID tasks have achieved great promotion in performance.
Extracting discriminative cues is critical for object Re-ID.
In the past few years, CNN-based methods have predominated the Re-ID tasks~\citep{pcb,wang2018person,li2018harmonious,suh2018part,chen2019abd,10054607}.
Early works extract robust features by applying deep CNNs.
For example, Wang~\emph{et al.}~\citep{wang2018person} cascade multiple convolutional layers to obtain different semantic information.
However, these methods are time-consuming.
On the other hand, some works~\citep{pcb,MGN,pit,alignedreid,Sun_2019_CVPR,Pyramid} split the feature maps into multiple parts along different directions for fine-grained information.
Sun \emph{et al.}~\citep{pcb} utilize horizontal partitions to learn discriminative local features.
Zhang \emph{et al.}~\citep{alignedreid} not only partition the feature maps for learning local cues but also design a dynamic alignment mechanism to measure the similarity between the same semantic parts.
Furthermore, Zheng~\emph{et al.}~\citep{Pyramid} divide deep feature maps into multi-scale sub-maps to incorporate local and global information.
However, these methods treat every part equally, which may lose some crucial information.
To address this issue, several attention-based methods~\citep{si2018dual,xu2018attention, 10319076} are employed to suppress irrelevant features and enhance discriminative ones.
In addition, Huang \emph{et al.}~\citep{10054607} design the graph attention mechanism to extract useful person representations.
Xu \emph{et al.}~\citep{xu2018attention} introduce pose estimation to guide attention generation.
Inspired by~\citep{nonlocal}, Zhang \emph{et al.}~\citep{zhang2020relation} generate attention maps by considering the pixel relation.
Although these methods have made great progresses in object Re-ID, CNN-based features generally focus on the local discriminative regions, which may cause over-fitting and ignore global important information.
To address these issues, in this work we introduce a novel pure Transformer framework for a more compact representation.
\subsection{Transformers for Object Re-Identification}
Due to its global modeling capabilities, Transformers become the mainstream models in the field of Natural Language Processing (NLP)~\citep{GPT}.
For vision tasks, Dosovitskiy \emph{et al.}~\citep{vit} propose a ViT model and achieve excellent performances.
%
%
Following ViT, some researchers~\citep{He_2021_ICCV,zhang2021hat,lai2021transformer,chen2021oh,chen2022structure} introduce Transformers into the Re-ID field for robust and discriminative global representations.
For example, He~\emph{et al.}~\citep{He_2021_ICCV} propose a pure Transformer model with the side information embeddings and designing a jigsaw patch module.
Zhang \emph{et al.}~\citep{zhang2021hat} introduce Transformer layers to hierarchically aggregate the multi-scale features from CNN-based backbones.
Gao \emph{et al.}~\citep{gao2024part} introduce part representation learning with a teacher-student decoder for occluded person Re-ID.
Lu \emph{et al.}~\citep{lu2023learning} propose a progressive modality-shared Transformer for effective visible-infrared person Re-ID.
Furthermore, Chen~\emph{et al.}~\citep{chen2022structure} utilize the key points and Transformers to extract the structure-aware information for visible-infrared person Re-ID.
However, these Transformer-based methods treat the class token as the feature representation, which may neglect the fine-grained cues among the patch tokens.
To address this issue, Zhu \emph{et al.}~\citep{AAformer} employ the optimal transport algorithm~\citep{cuturi2013sinkhorn} to define the semantics of patch tokens and use local Transformers for local-aware representations.
However, the semantics defined by the auxiliary algorithms may interfere with the complex background and mislead the feature representations.
On the other hand, Wang \emph{et al.}~\citep{wang2022pose} split the feature maps horizontally and align the patch tokens with the pose information for better local features.
Although good performances are achieved, the introduced key-point algorithm~\citep{HRnet} is not reliable in many complex scenes, and they may neglect the spatial relation among the patch tokens.
Recently, Yan \emph{et al.}~\citep{yan2023learning} integrate the advantages of CNNs and Transformers, and propose a convolutional multi-level Transformer for local-aware object Re-ID.
Wang \emph{et al.}~\citep{wang2024top} propose the token permutation for multi-spectral object Re-ID. 
Zhang \emph{et al.}~\citep{zhang2024magic} select diverse tokens from Transformers for multi-modal object Re-ID.
Furthermore, Liu \emph{et al.}~\citep{liu2021video} propose a trigeminal Transformers for video-based person Re-ID.
Liu \emph{et al.}~\citep{liu2023deeply} propose deeply coupled convolution-Transformer with spatial--temporal complementary learning for video-based person Re-ID.
Inspired by the great ability of vision-language models, Yu \emph{et al.}~\citep{yu2024tf} propose a text-free CLIP model for video-based person Re-ID.
Different from previous works, we propose a local multi-layer fusion module to further mine the spatial information and directly extract the fine-grained cues from patch tokens.
\begin{figure*}[htbp]
    \centering
    \includegraphics[width=0.9\linewidth]{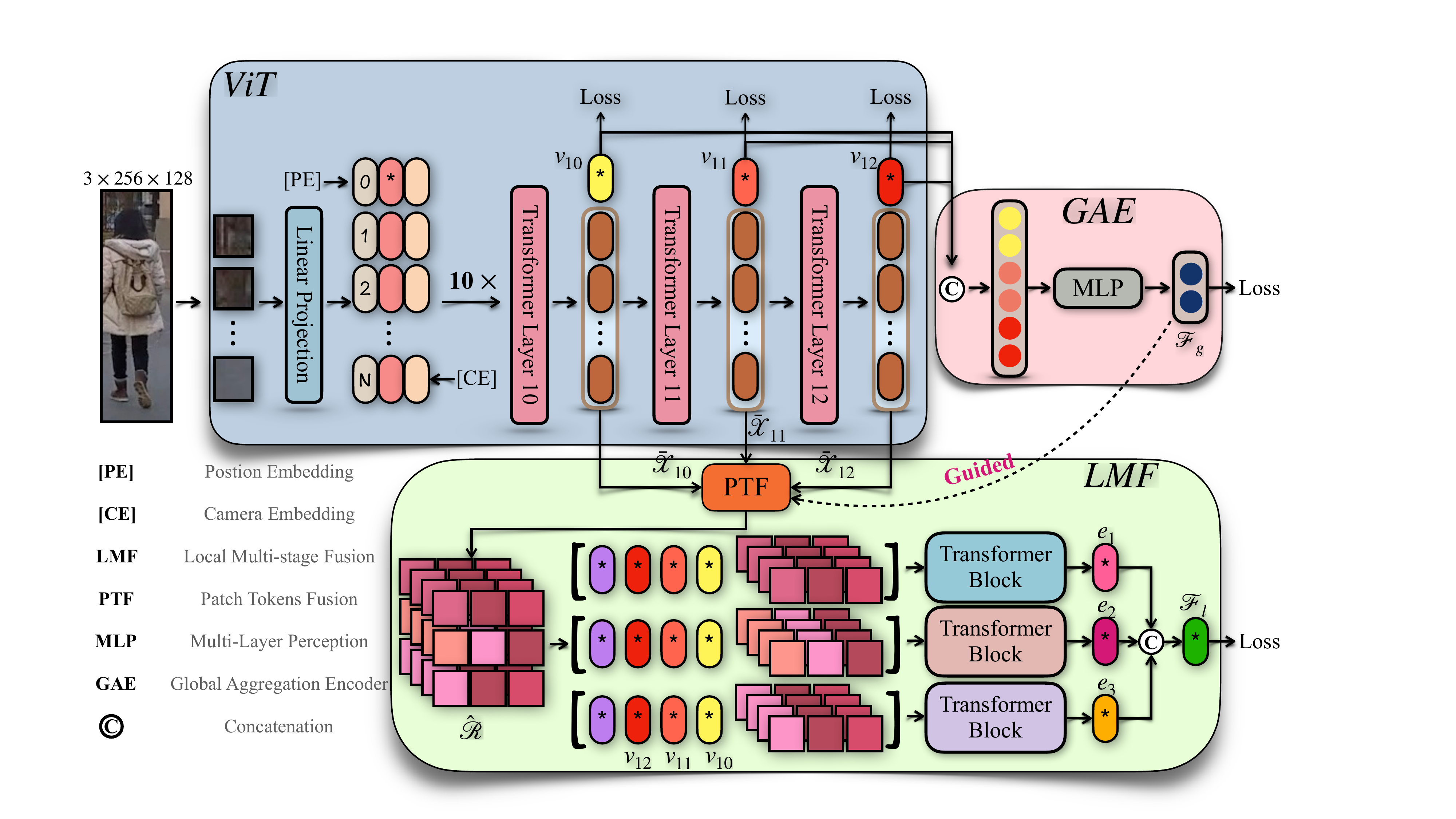}
    \caption{Our proposed GLTrans. The Vision Transformer (ViT) with side information embedding (cameras or viewpoints) is adopted as the backbone to obtain multi-layer class tokens and patch tokens. Then, two branches are used to extract global and local representations.
    The Global Aggregation Encoder (GAE) generates global representations by incorporating multi-layer class tokens, while Local Multi-layer Fusion (LMF) takes patch tokens as inputs to further extract the local-wise discriminative features.}
    \label{fig:framework}
\end{figure*}
\section{Proposed Method}
As illustrated in Fig.~\ref{fig:framework}, the proposed framework (i.e., GLTrans) mainly includes three key modules: Vision Transformer (ViT), Global Aggregation Encoder (GAE) and Local Multi-layer Fusion (LMF).
We will elaborate on these key components in the following subsections.
\subsection{Revisiting Vision Transformer}
To begin with, we briefly review the ViT~\citep{vit}.
Given an image $x\in \mathbb{R}^{H\times W\times C}$, where $H$, $W$ and $C$ represent the height, width and channel, respectively, we utilize a sliding window to obtain overlapped image patches.
Then, the image is disassembled into $N$ patches ($N=\lfloor \frac{H+S-P}{S} \rfloor \times \lfloor \frac{W+S-P}{S} \rfloor$).
Specifically, we set the step size as $S=12$ and the patch size as $P=16$.
Each patch is linearly projected into a $D$-dimensional vector.
In addition, a class token $v_{1}$ is prefixed to these verctors.
As a result, the vector sequence is obtained by concatenating these vectors, as follows:
\begin{equation}
    \mathcal{X}_1 = [v_{1}; \phi(x_{1}^1);\cdots;\phi(x_{1}^N)] + \mathcal{P} + \mathcal{S},
\end{equation}
where  $\mathcal{P} \in \mathbb{R}^{(N+1)\times D}$ and $\mathcal{S} \in \mathbb{R}^{(N+1)\times D}$ are learnable embeddings, representing the position information and side information (cameras or viewpoints), respectively.
\subsection{Global Aggregation Encoder}
Many Transformer-based methods~\citep{vit,Zhu_2022_CVPR} have shown that using the class token from the last layer achieves attractive results.
However, as depicted in Fig.~\ref{fig:multiple layer}, the last few layers of the ViT also present strong representational abilities and the best representations are not always in the last layer.
Therefore, utilizing the last class token may be a suboptimal strategy.
To address this, we propose a GAE module.
Specifically, we collect class tokens from the last three layers to construct a global aggregated vector $\hat{\mathcal{F}}_g$,
\begin{equation}
    \hat{\mathcal{F}}_g = [v_{10},v_{11},v_{12}].
\end{equation}
For generating the useful global information and reducing the irrelevant cues, we directly feed the vector $\hat{\mathcal{F}}_g$ into a Fully-Connected layer (FC) and a GeLU activation function~\citep{GeLU} to obtain a comprehensive global representations,
\begin{equation}
  \mathcal{F}_g = GeLU(FC(\hat{\mathcal{F}}_g)).
\end{equation}

As shown in Fig.~\ref{fig:multiple layer}, the features from the last few layers exhibit strong discriminative and diverse semantics.
However, simply concatenating or adding these features may enhance some erroneous information, as depicted in Fig.~\ref{fig:multiple layer} (a) and Fig.~\ref{fig:multiple layer} (f).
Therefore, we introduce a FC layer to recognize the relationships among them and further highlight the useful ones.
The experiments in Sec. 4.4 demonstrate that this simple structure can efficiently extract comprehensive features.
\subsection{Local Multi-layer Fusion}
There are some pure Transformer-based Re-ID methods to explore global and local representations~\citep{song2023boosting,AAformer}.
However, most of them extract the global and local representations independently, which may neglect that the local-global interaction can complement each other.
On the other hand, most of previous methods~\citep{vit,He_2021_ICCV} adopt patch tokens from the last layer, which inevitably neglect diverse semantics from different layers.
To address these issues, we propose the Local Multi-layer Fusion (LMF), which contains Patch Token Fusion (PTF), Global-guided Multi-head Attention (GMA) and Part-based Transformer Layers (PTL).
They are described as follows.
\subsubsection{Patch Token Fusion}
As shown in Fig.~\ref{fig:multiple layer}, features from different Transformer layers have diverse semantics.
They can complement each other.
In addition, the cascade Transformer blocks neglect the spatial information among the patch tokens.
Thus, we propose the PTF to obtain a compact local representation by aggregating multi-layer patch tokens and enhancing these spatial relationships.

\begin{figure}[htbp]
    \centering
    \includegraphics[height=0.9\linewidth]{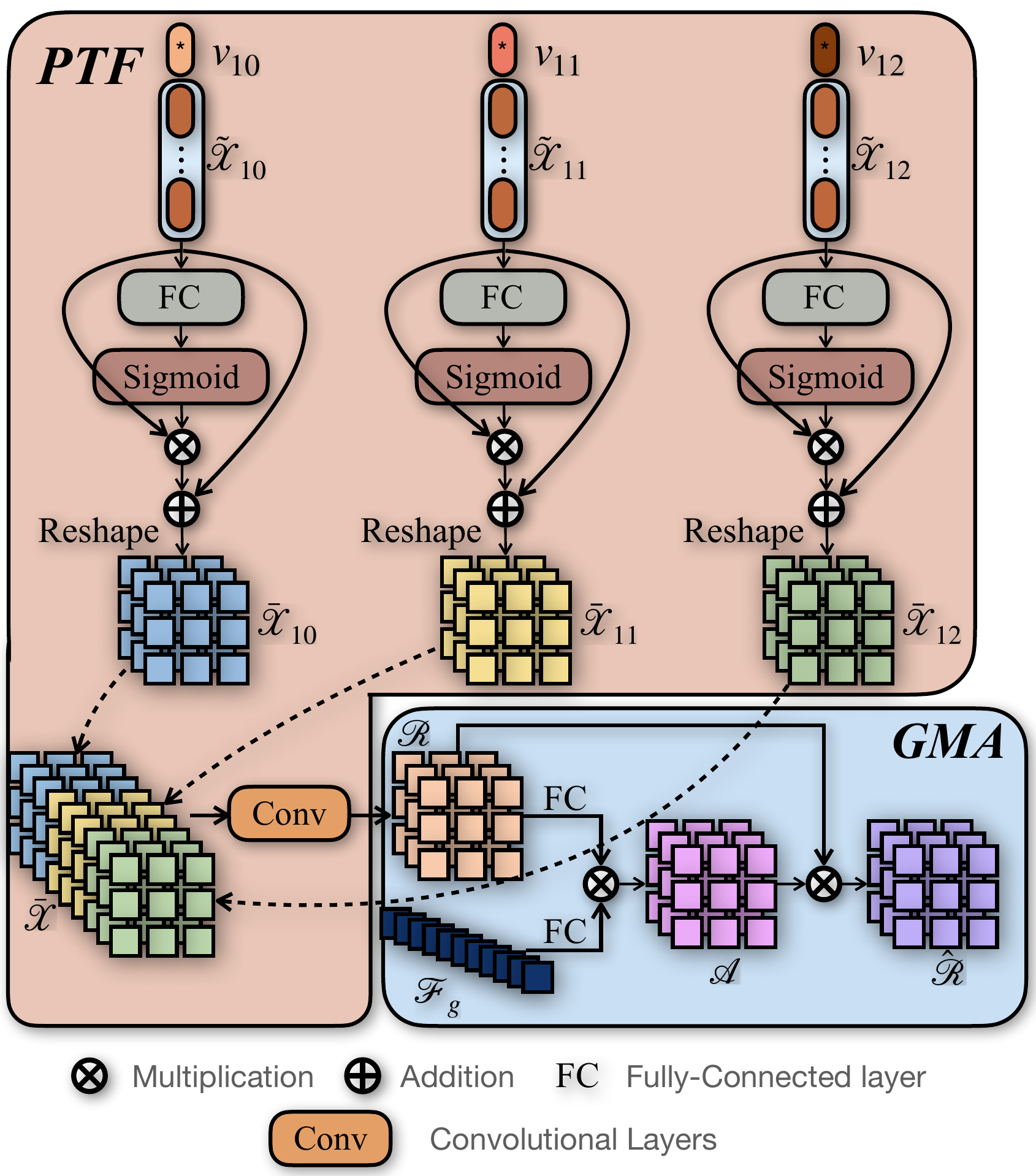}
    \caption{Our proposed Patch Token Fusion (PTF) and Global-guided Multi-head Attention (GMA).}
    \label{fig:PTF}
\end{figure}

Specifically, as shown in Fig.~\ref{fig:PTF}, we take the intermediate features $\mathcal{X}_{10},\mathcal{X}_{11},\mathcal{X}_{12}$ and split them into patch tokens $\tilde{\mathcal{X}_{l}}\in \mathbb{R}^{N\times D}$ and class token $v_{l} \in \mathbb{R}^{1 \times D}$.
Then, the patch tokens $\tilde{\mathcal{X}_{l}}$ are fed into two linear transformations followed by Sigmoid and GeLU activation functions to generate the weight mask $\mathcal{S}_{l}$,
\begin{equation}
    \mathcal{S}_{l} = {Sigmoid}\left({GeLU}(\tilde{\mathcal{X}_{l}}\times \mathcal{W}^{1}_{l})\times \mathcal{W}^{2}_{l}\right),
\end{equation}
where $\mathcal{W}^{1}_{l} \in \mathbb{R}^{D\times \frac{D}{r_1}}$ and $\mathcal{W}^{2}_{l} \in \mathbb{R}^{\frac{D}{r_1}\times D}$ are learnable parameters.
$r_1$ is a reduction ratio of feature dimensions.
Furthermore, we obtain enhanced patch tokens $\bar{\mathcal{X}}_{l}$ by
\begin{equation}
    \bar{\mathcal{X}}_{l} = \mathcal{S}_l \otimes \tilde{\mathcal{X}_{l}}+\tilde{\mathcal{X}_{l}},
\end{equation}
where $\mathcal{S}_l\in \mathbb{R}^{N\times D}$.
Afterwards, we reshape enhanced patch tokens $\bar{\mathcal{X}}_{l}$ according their spatial position and concatenate them,
\begin{equation}
    \bar{\mathcal{X}} = {Concat}[\bar{\mathcal{X}}_{10}; \bar{\mathcal{X}}_{11}; \bar{\mathcal{X}}_{12}],
\end{equation}
where $\bar{\mathcal{X}}\in \mathbb{R}^{\hat{H}\times\hat{W}\times 3D} (N=\hat{H}\times\hat{W})$.
%
%
Inspired by previous works~\citep{song2023boosting,peng2021conformer}, we also apply $1\times 1$ and $3\times3$ convolutions following a BN and GeLU activation function to explore the spatial relation among the patch tokens,
\begin{equation}
    \mathcal{R} = {GeLU}\left({BN}\left({Conv}(\bar{\mathcal{X}})\right)\right),
\end{equation}
where $\mathcal{R}\in \mathbb{R}^{\hat{H}\times\hat{W}\times D}$ is the fused patch tokens of PTF.
\subsubsection{Global-guided Multi-head Attention}
Although the global feature $\mathcal{F}_g$ may neglect some fine-grained information, it contains rich semantic information~\citep{vit}.
Considering this fact, we design a guidance mechanism to further boost the discriminative representation of the enhanced patch tokens.
Specifically, as shown in the right-bottom of Fig~\ref{fig:PTF}, we treat the global feature $\mathcal{F}_g\in \mathbb{R}^{O\times1\times D'}$ as query.
Meanwhile, the enhanced patch tokens $\mathcal{R}$ reshaped to $\mathcal{R}^{(r)}\in \mathbb{R}^{O\times N \times D'}$ are treated as key.
Here, $O$ is the number of heads and $N=O\times D'$.
We set the number of heads $O$ to be 12.
Formally, the Global-guided Multi-head Attention (GMA) can be expressed as:
\begin{equation}
    \mathcal{A}_p = {Sigmoid}\left((\mathcal{F}_g\times \mathcal{W}_{q})\otimes(\mathcal{R}^{(r)}_p \times \mathcal{W}_{k})\right),
\end{equation}
where  $\mathcal{A}_p\in \mathbb{R}^{O\times1\times D'}$ is the score of \emph{p-th} patch token $\mathcal{R}_p$.
$\mathcal{W}_{k}\in\mathbb{R}^{O\times D'\times \frac{D'}{r_2}}$ and $\mathcal{W}_{q}\in\mathbb{R}^{O \times D'\times \frac{D'}{r_2}}$ are learnable parameters.
$r_2$ is a reduction ratio.
Then, we can obtain the attention map $\mathcal{A} \in \mathbb{R}^{O\times N\times D'}$ by concatenating the attention scores,
\begin{equation}
    \mathcal{A} = \left[\mathcal{A}_1, \mathcal{A}_2, \cdots, \mathcal{A}_{N}\right].
\end{equation}
Finally, the resulted patch tokens $\hat{\mathcal{R}}$ are defined as:
\begin{equation}
    \hat{\mathcal{R}} = \mathcal{R} + \mathcal{R} \times \mathcal{A}.
\end{equation}
From the above procedure (especially Eq.~8), one can see that the patch tokens are further enhanced by the the global feature $\mathcal{F}_g$.
As a result, they can deliver more discriminative features.
\subsubsection{Part-based Transformer Layers}
Recently, some part-based Transformer methods~\citep{AAformer,zhang2021hat,yan2023learning} attempt to extract the local fine-grained information by dividing the feature maps and directly supervising these partition features.
However, they may neglect two issues: 1) The patch tokens in Transformers contain rich semantics for global modeling.
Thus, it may introduce irrelevant information by directly utilizing the partition features for training and testing.
2) The interaction between the local-wise patch tokens and global features can enhance the robustness of local representations.
However, previous works neglect to incorporate global information to obtain comprehensive representations.
For example, PTCR~\citep{li2022pyramidal} utilizes a token perception module to extract local discriminative cues and leverages the powerful PVTv2~\citep{wang2022pvt} backbone to achieve impressive performance.
Based on these facts, we present Part-based Transformer Layers (PTL) to extract discriminative local features.
Our method goes beyond solely extracting information from the outputs of the last few Transformer layers.
We incorporate the global information during the extraction of local representations, which provides additional references.
The proposed method can capture both local and global cues, leading to more comprehensive and informative representations for object Re-ID.

Specifically, as shown in the bottom of Fig.~\ref{fig:framework}, we split the enhanced patch tokens $\hat{\mathcal{R}}$ into three horizontal stripes and construct three local sequences.
In addition, we prefix the last three class tokens $v_{10},v_{11},v_{12}$ to each part sequence for further utilizing the global information.
To improve the feature representation ability, we introduce an additional class token $e_{t}(t=1,2,3)$ to each local sequence.
The input for part-based transformers is formulated as:
\begin{equation}
      \breve{\mathcal{R}_t} = \left[e_{t},v_{10},v_{11},v_{12},x^{\frac{(t-1)\times N}{3}+1 }_t,\cdots,x_t^{\frac{t\times N}{3}+1}\right],
\end{equation}
where $x_t^{n}$ is the $n$-\emph{th} patch token in the $t$-\emph{th} part.
Then, each part-wise sequence $\breve{\mathcal{R}_t}$ is followed by two independent Transformer layers, which contain a Multi-head Self-attention Layer (MSA), a Feed-Forward Network (FFN), Layer Normalizations (LN)~\citep{ba2016layer} and residual connections,
\begin{equation}
  \breve{\mathcal{R}}'_t = \breve{\mathcal{R}_t} + {MSA} \left({LN}({\breve{R}_t})\right),
\end{equation}
\begin{equation}
  \breve{\mathcal{R}_t}''_t = \breve{\mathcal{R}_t}' + {FFN}\left({LN}({\breve{R}_t}')\right).
\end{equation}
Finally, to encourage the PTL to learn more diverse and complementary information, we supervise the concatenated class tokens $\mathcal{F}_l=[e_{1}, e_{2}, e_{3}]$ with loss functions.

With the above modules, we not only obtain local representations by dividing the feature map into local parts, but also introduce global information through multiple global class tokens ($v_{10},v_{11},v_{12}$) into the part-based self-attention mechanism.
Thus, both global and local cues are aggregated into class tokens, resulting in comprehensive feature representations.
\begin{table*}[htbp]
    \caption{Quantitative comparison of state-of-the-art methods on three public person Re-ID datasets.}
    \centering
    \begin{tabular}{lc|cccccc}
    \hline
    \multirow{2}{*}{Method}&
    \multirow{2}{*}{Backbone}&
    \multicolumn{2}{c}{Market1501}&
    \multicolumn{2}{c}{MSMT17}&
    \multicolumn{2}{c}{DukeMTMC}\\
                &             &mAP         &Rank1         & mAP         & Rank1         & mAP         & Rank1\\
    \hline
    SPReID~\citep{kalayeh2018human}      &ResNet152     &83.4        &93.7          &--           &--             &73.3         &85.9\\
    CASN~\citep{zheng2019re}        &ResNet50     &82.8        &94.4          &--           &--             &73.7         &87.7\\
    BATNet~\citep{fang2019bilinear}      &ResNet50     &84.7        &95.1          &56.8         &79.5           &77.3         &87.7\\
    MGN~\citep{MGN}         &ResNet50     &86.9        &95.7          &--           &--             &78.4         &88.7\\
    ABDNet~\citep{chen2019abd}      &ResNet50     &88.3        &95.6          &60.8         &82.3           &78.6         &89.0\\
    Pyramid~\citep{Pyramid}     &ResNet101    &88.2        &95.7          &--           &--             &79.0         &89.0\\
    OSNet~\citep{zhou2019omni}       &OSNet        &84.9        &94.8          &52.9         &78.7           &73.5         &88.6\\
    SNR~\citep{jin2020style}         &ResNet50     &84.7        &94.4          &--           &--             &73.0         &85.9\\
    RGA-SC~\citep{zhang2020relation}      &ResNet50     &88.4        &\textbf{96.1}          &57.5         &80.3           &--           &--\\
    SCSN~\citep{SCSN}        &ResNet50     &88.5        &95.7          & --          & --            &79.0         &\textbf{91.0}\\
    CDNet~\citep{CDnet} &CDNet  &86.0   &95.1  &54.7  &78.9  &76.8  &88.6\\
    PAT~\citep{Li_2021_CVPR}  &ResNet50   &88.0   &95.4   &--
    &--  &78.2  &88.8\\
    ISP~\citep{ISP}         &HRNet48      &88.6        &95.3          & --          & --            &80.0         &89.6\\
    FED~\citep{wang2022feature}        &ResNet50     &86.3        &95.0          & --          & --            &78.0         &89.4\\
    Nformer~\citep{wang2022nformer}     &ResNet50       &\textbf{91.1}        &94.7          &59.8         &77.3           &\textbf{83.5}         &89.4\\
    \hline
    AAformer~\citep{AAformer}    &ViT-B/16    &87.7        &95.4          &62.6         &{83.1}             &80.0         &90.1\\
    TransReID~\citep{He_2021_ICCV}   &ViT-B/16     &88.9        &95.2          &\underline{67.4}         &\underline{85.3}           &82.0         &\textbf{90.7}\\
    APD~\citep{lai2021transformer}         &ResNet50     &87.5        &95.5          &57.1         &79.8           &74.2         &87.1\\
    HAT~\citep{zhang2021hat}         &ResNet50     &89.8        &\underline{95.8}          &61.2         &82.3           &81.4         &{90.4}\\
    ADSO~\citep{Zhang_2021_CVPR}  &ResNet50   &87.7   &94.8   &--  &--  &74.9  &87.4\\
    PFD~\citep{wang2022pose}         &ViT-B/16    &{89.6}        &95.5          &{65.1}         &82.7           &{82.2}         &\underline{90.6}\\
    DCAL~\citep{Zhu_2022_CVPR}        &ViT-B/16    &87.5        &94.7          &64.0         &{83.1}           &80.1         &89.0\\
    \hline
    GLTrans (Ours)        &ViT-B/16    &\underline{90.0}        &{95.6}         &\textbf{69.0}       &\textbf{85.8}         &\underline{82.4}         &\textbf{90.7}\\
    \hline
    \end{tabular}
    \label{table:person sota}
\end{table*}
\subsection{Loss Functions}
To train our proposed framework, we follow previous works and adopt the cross-entropy loss and triplet loss~\citep{tripletloss}.
Specifically, the cross-entropy loss is defined as
\begin{equation}
        \mathcal{L}_{c} = - \frac{1}{\mathcal{P}\times \mathcal{K}} \sum^\mathcal{P}_{i=1}\sum^\mathcal{K}_{j=1}{log}\frac{{exp}(W^T_{y_i, j} \cdot f_{i, j})}{\sum_{c=1}^{\mathcal{P}\cdot \mathcal{K}}{exp}(W_c^T \cdot f_{i, j})},
\end{equation}
where $\mathcal{P}$ and $\mathcal{K}$ are the number of identities and sampled images of each identity.
$W^T_{y_i, j}$ means the weight parameters of the $i$-\emph{th} label in the classification layer.
$f_{i, j}$ represent the feature corresponding to the $j$-\emph{th} sample of the label $i$.
In addition, the triplet loss is formulated as,
\begin{equation}
  \mathcal{L}_t = {log}\left[1+exp(||f_a-f_p||^2_2-||f_a-f_n||^2_2)\right],
\end{equation}
where $f_a$, $f_p$, and $f_n$ are the anchor feature, positive features and negative features, respectively.
Finally, the global feature $\mathcal{F}_g$, the local feature $\mathcal{F}_l$ and the class token embbeding $\mathcal{F}_{cls}$ are trained with $L_c$ and $L_t$, while the class token $v_{10}$ and $v_{11}$ trained with $L_c$.
The overall loss is computed as follow:
\begin{equation}
    \mathcal{L}_{total} = \frac{1}{3} \sum_{u\in\{\mathcal{F}_g, \mathcal{F}_l, \mathcal{F}_{cls}\}} \left(\mathcal{L}_c(u) +\mathcal{L}_t(u)\right) + \frac{1}{2} \sum_{z \in\{v_{10}, v_{11}\}} \left(\mathcal{L}_c(z)\right).
\end{equation}
During testing, we concatenate global features, local features, and the last class token embbeding as the image representation,
\begin{equation}
    \mathcal{F} = [\mathcal{F}_g, \mathcal{F}_l,\mathcal{F}_{cls}],
\end{equation}
where $\mathcal{F}_{cls}$ is the last class token embbeding $v_{12}$.
\begin{table}[htbp]
      \centering
      \caption{Statistics of used datasets.}
      \resizebox{0.5\textwidth}{!}{
      \begin{tabular}{p{28mm}<{\centering}|p{9mm}<{\centering}p{6mm}<{\centering}p{10mm}<{\centering}|p{12mm}<{\centering}}
          \hline
           Dataset           &Object    &ID    &Image  &Cam(View)\\
           \hline
           MSMT17            &Person    &4,101  &126,441   &15\\
           Market1501       &Person    &1,501  &32,668    &6\\
           DukeMTMC-ReID     &Person    &1,404  &36,441    &8\\
           VeRi-776          &Vehicle   &776    &49,357    &20(8)\\
           \hline
      \end{tabular}
      }
      \label{table:datasets}
      \vspace{-6mm}
  \end{table}
\section{Experiments}
\subsection{Datasets and Evaluation Metrics}
To fully verify the effectiveness of our proposed framework, we perform experiments on four large-scale object Re-ID datasets, \emph{i.e.}, Market1501~\citep{market1501}, DukeMTMC-ReID~\citep{Duke}, MSMT17~\citep{MSMT17} and VeRi-776~\citep{VeRi}.
The details of these datasets are summarized in Tab.~\ref{table:datasets}.
Following previous Re-ID works, we adopt mean Average Precision (mAP) and Cumulative Matching Characteristics (CMC) at Rank1 as our evaluation metrics.
%
\subsection{Implementation Details}
We implement our framework based on the PyTorch toolbox.
Experimental devices include an Intel(R) Xeon(R) Platinum 8350C CPU and one NVIDIA GTX 3090 GPU (24G memory).
For model training, we uniformly resize all person images to $256 \times 128$ and all vehicle images to $256\times 256$, then followed by random cropping, horizontal flipping and random erasing~\citep{randomerasing} as data augmentations.
In addition, there are $\mathcal{B} = \mathcal{P} \times \mathcal{K}$ images sampled to the triplet loss and cross-entropy loss in a mini-batch for every training iteration.
We randomly select $\mathcal{P} = 16$ identities and $\mathcal{K} = 4$ images for each identity.
We employ SGD as our optimizer for total 200 epochs with a momentum of $0.9$ and the weight decay of $1\times e^{-4}$.
The initial learning rate is $ 8 \times 10^{-3} $ with a cosine learning rate decay.
We adopt the ViT-B/16~\citep{vit} with a stride 12 as our backbone, which is pre-trained on ImageNet-21k~\citep{imagenet_cvpr09} and then fine-tuned on ImageNet-1k~\citep{imagenet_cvpr09}.
We will release the source code for reproduction.
\begin{table}[htbp]
      \caption{Quantitative comparison of state-of-the-art methods on VeRi-776.}
    \centering
    \resizebox{0.5\textwidth}{!}{
    \begin{tabular}{lc|cc}
    \hline
    \multirow{2}{*}{Method}&
    \multirow{2}{*}{Backbone}&
    \multicolumn{2}{c}{VeRi-776}\\
                &             &mAP         &Rank1\\
    \hline
    PPReID~\citep{he2019part}      &ResNet50     &72.5        &93.3\\
    SAN~\citep{qian2020stripe}         &ResNet50     &72.5        &93.3\\
    UMTS~\citep{jin2020uncertainty}        &ResNet50     &75.9        &95.8\\
    VANet~\citep{chu2019vehicle}       &ResNet50     &66.3        &95.8\\
    SPAN~\citep{chen2020orientation}        &ResNet50     &68.9        &94.0\\
    PGAN~\citep{zhang2020part}        &ResNet50     &79.3        &96.5\\
    PVEN~\citep{meng2020parsing}        &ResNet50     &79.5        &95.6\\
    SAVER~\citep{khorramshahi2020devil}       &ResNet50     &79.6        &96.4\\
    CFVMNet~\citep{sun2020cfvmnet}     &ResNet50     &77.1        &95.3\\
    GLAMOR~\citep{suprem2020looking}      &ResNet50     &80.3        &96.5\\
        MPC~\citep{li2021exploiting}  &ResNet50   &{80.9}  &96.2\\
    MsKAT~\citep{li2022mskat}     &ResNet50   &\underline{82.0}  &\underline{97.1}\\
    TransReID~\citep{He_2021_ICCV}   &ViT-B/16    &\underline{82.0}       &\underline{97.1}\\
    DCAL~\citep{Zhu_2022_CVPR}        &ViT-B/16     &80.2        &{96.9}\\
    \hline
    GLTrans (Ours)        &ViT-B/16     &\textbf{82.9}        &\textbf{97.5}\\
    \hline
    \end{tabular}
    }
    \label{tabel:vehicle sota}
\end{table}
\subsection{Comparison with State-of-the-art Methods}
In Tab.~\ref{table:person sota} and Tab.~\ref{tabel:vehicle sota}, our GLTrans is compared with other state-of-the-art methods on four benchmarks.

\textbf{Market1501}:
Tab.~\ref{table:person sota} shows the performances of compared methods on Market1501.
Comparing with other models, we can see that our framework achieves very competitive results, especially in mAP.
While the Rank1 score of our model is inferior to some compared methods, (\emph{e.g.}, ISP~\citep{ISP}, HAT~\citep{zhang2021hat} and SCSN~\citep{SCSN}).

\textbf{MSMT17}:
As shown in Tab.~\ref{table:person sota}, our model achieves the best performance in terms of mAP and Rank1 on MSMT17.
It is worth pointing out that the mAP score of our model is higher than ISP~\citep{ISP} and TransReID~\citep{He_2021_ICCV} by $1.4\%$ and $1.1\%$, respectively.
The results indicate that the fusion of multi-layer features guided by global cues can obtain complementary and fine-grained feature representations.

\textbf{DukeMTMC-ReID}:
Tab.~\ref{table:person sota} shows that our method achieves very comparable performances in terms of mAP and Rank1 on DukeMTMC-ReID.
More specifically, our GLTrans surpasses TransReID~\citep{He_2021_ICCV}, AAformer~\citep{AAformer} and PFD~\citep{wang2022pose} by $0.4\%$, $2.4\%$, and $0.2\%$ on mAP score, respectively.
It indicates that by exploring the complementary local and global information, our GLTrans can obtain more robust representations.

\textbf{VeRi-776}:
To further verify the ability of our proposed model, we also compare GLTrans with some vehicle Re-ID models~\citep{He_2021_ICCV, Zhu_2022_CVPR}.
As shown in Tab.~\ref{tabel:vehicle sota}, our framework achieves the best performance in mAP and Rank1.
Specifically, our GLTrans outperforms TransReID~\citep{He_2021_ICCV} by $0.9\%$ and $0.4\%$ in mAP/Rank1.
In fact, vehicle Re-ID faces small inter-class variances and significant intra-class variances.
This implies that recognizing local information has a significant impact on discriminative feature representations.
The results in Tab.~\ref{tabel:vehicle sota} show that, unlike other ViT-based methods, our proposed approach considers both local and global cues, resulting in the excellent performance.
\subsection{Ablation Study}
In this subsection, we conduct ablation experiments to verify the effect of the key components.
The baseline method adopts a ViT-B/16~\citep{vit} with overlapped patches and camera/viewpoint information embedding.
%
%
All the ablation studies are conducted on the MSMT17 dataset.
However, the results on other datasets show similar trends.

\textbf{Comparisons with other part-based methods.}
We verify the effect of different part-based methods.
Here, PCB$^*$ refers to the reproduction of PCB~\citep{pcb} based on ViT.
Specifically, we reshape the output tokens of ViT into a feature map based on their spatial coordinates and extract local features by applying the average pooling to the divided feature map.
During testing, we concatenate the averaged features with class tokens.
MSF* indicates that we aggregate tokens from multiple layers using GAE and PTF modules to obtain global and local representations.
The local representations are obtained by applying the average pooling to the fused feature map.

\begin{figure*}[htbp]
    \centering
    \includegraphics[width=1\linewidth]{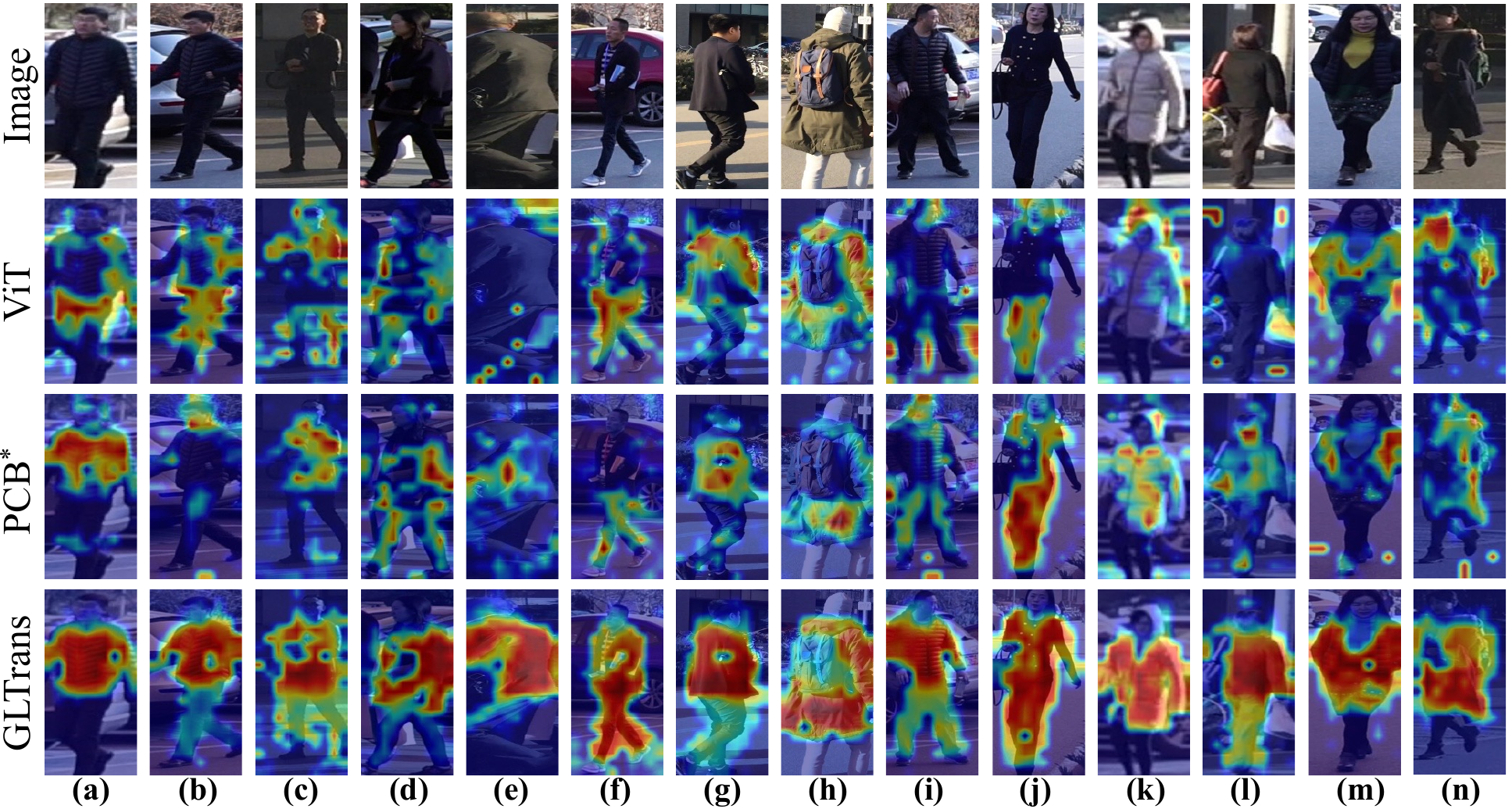}
    \caption{Visualization of the differences between ViT, PCB$^*$ and GLTrans by Grad-CAM~\citep{gram_cam}.
    Deeper red colors signify higher weights. The first row is the input images. The second, third and fourth rows are the activation maps produced by ViT, PCB$^{*}$ and our GLTrans, respectively.}
    \label{fig:heatmap}
\end{figure*}

As shown in Tab.~\ref{tab:part-based models}, compared with the baseline, it clearly achieves improvements by using the part-aware method and multi-layer fusion.
The results indicate that the patch tokens contain valuable fine-grained information.
In addition, the outputs of the last few layers of ViT complement each other.
Although PCB$^*$ can explore local cues from patch tokens, it can not fully explore complementary local cues by solely using the outputs of the last layer.
Fig.~\ref{fig:multiple layer} and Fig.~\ref{fig:heatmap} clearly show these facts.
The results shown in the last row of Tab.~\ref{tab:part-based models} illustrate that simultaneously utilizing the part modeling methods and multi-stage fusion mechanisms achieve best performances.
\begin{table}[hbtp]
    \centering
    \caption{Performance comparison with different part-based methods.}
    \begin{tabular}{c|cc}
    \hline
             Model  &mAP &Rank1\\
    \hline
      Baseline        &66.3    &84.0\\
      PCB$^{*}$       &68.2    &85.2\\
      MSF$^{*}$       &67.6    &84.8\\
    \hline
      Ours            &\textbf{69.0}    &\textbf{85.8}\\
    \hline
    \end{tabular}
    \label{tab:part-based models}
\end{table}

\textbf{The impacts of different components.}
In Tab.~\ref{tab:component study}, we present the ablation studies of GAE and LMF modules.
Model-1 corresponds to the baseline.
Compared with Model-2, we can see that employing the GAE significantly improves the performance over the baseline by $1.5\%$ on mAP and $0.8\%$ on Rank1 accuracy.
The main reason is that the GAE can aggregate several class tokens, representing multiple global semantic information of input images.
Model-3 means that we add the PTL to obtain the local information.
One can see that the performance further improves by $0.7\%$ and $0.3\%$ in terms of mAP and Rank1, respectively.
The main reason is that our PTL makes the model obtain a more robust representation by focusing on more detailed information.
Model-4 and Model-5 mean that we further introduce the PTF and GMA.
As shown in Tab.~\ref{tab:component study}, the final model further provides $0.5\%$ mAP and $0.3\%$ Rank1 accuracy improvements, respectively.
It indicates that aggregating and enhancing the multi-layer patch tokens can explore more discriminative and robust features.
The above results clearly demonstrate the effectiveness of our proposed modules.
%
\begin{table}[hbtp]
    \centering
    \caption{Performance comparison of key components.}
    \resizebox{0.5\textwidth}{!}{
    \begin{tabular}{c|ccccc|cc}
    \hline
             \multirow{2}{*}{Model} &\multirow{2}{*}{Baseline}  &\multirow{2}{*}{GAE}   &\multicolumn{3}{c|}{LMF}  &\multicolumn{2}{c}{MSMT17}\\
         &&  &PTL &PTF &GMA &mAP &Rank1\\
    \hline
     1&\checkmark &$\times$ &$\times$ &$\times$ &$\times$      &66.3  &84.0\\
     2&\checkmark &\checkmark &$\times$  &$\times$  &$\times$   &67.8  &85.2\\
     3&\checkmark &\checkmark &\checkmark &$\times$   &$\times$ &68.5  &85.5\\
     4&\checkmark &\checkmark &\checkmark &\checkmark   &$\times$ &68.8  &85.7\\
     5&\checkmark &\checkmark &\checkmark &\checkmark   &\checkmark &\textbf{69.0}  &\textbf{85.8}\\
    \hline
    \end{tabular}
    }
    \label{tab:component study}
\end{table}
\begin{table}[hbtp]
    \centering
    \caption{Performance comparison with different aggregated layers.}
    \begin{tabular}{c|cc}
    \hline
             Layer  &mAP &Rank1\\
    \hline
      9,10,11,12   &68.2    &85.2\\
      10,11,12     &\textbf{68.6}  &\textbf{85.8}\\
      11, 12       &67.8    &85.1\\
      12           &67.5    &85.1\\
    \hline
    \end{tabular}
    \label{tab:stage study}
\end{table}

\textbf{The impacts of aggregating different layers.}
The choice of which layers to aggregate is an important hyper-parameter in our framework.
Therefore, we investigate the impacts of aggregating different layers.
As shown in Tab.~\ref{tab:stage study}, we find that the highest performance is achieved by aggregating the last three layers of ViT.
It is verified that the features from the last few layers actually contain different semantic information.
They can complement and enhance each other.

\textbf{The impacts of different aggregation strategies in GAE.}
We exhibit experimental results of four aggregation strategies in Tab.~\ref{tab:fusion study}.
We first adopt addition and concatenation to aggregate the features, respectively.
As shown in the 2-\emph{th} and 3-\emph{th} rows, the performances are not satisfactory.
The main reason is that multiple features have different semantics.
This kind of straightforward methods can not explore the information among them and even destroy the feature distributions.
Meanwhile, in the 4-\emph{th} and 5-\emph{th} rows, we illustrate the results of employing different FC layers.
As can be observed, the best performance is achieved by using only one FC layer.
We argue that the FC layer can consider the relation between the multiple features with different semantics.
One FC layer is more easily trained and avoids over-fitting.
\begin{table}[hbtp]
    \centering
    \caption{Performance comparison with different aggregation strategies.}
    \begin{tabular}{c|cc}
    \hline
             Stage  &mAP &Rank1\\
    \hline
      Addition          &66.4    &84.7\\
      Concatenation     &66.6    &84.5\\
      One FC        &\textbf{69.0}    &\textbf{85.8}\\
      Two FC        &66.3    &84.1\\
    \hline
    \end{tabular}
    \label{tab:fusion study}
\end{table}

\textbf{The impacts of different partition numbers.}
It is of great interest to determine the optimal partition strategy for the feature map to extract local fine-grained features.
To this end, we conduct experiments to assess the impacts of varying the number of parts.
As depicted in Tab.~\ref{tab:chunking number}, dividing the feature map into more parts does not lead to performance improvement.
For instance, when the feature map is divided into six parts, the performance is comparable to that of three parts.
Consequently, in this paper, to keep the computational efficiency, we divide the feature map into three parts.
\begin{table}[hbtp]
    \centering
    \caption{Performance comparison with different partition numbers.}
    \begin{tabular}{c|cc}
    \hline
             Number  &mAP &Rank1\\
    \hline
      1          &68.3    &85.2\\
      2          &69.0    &85.7\\
      3          &\textbf{69.0}    &\textbf{85.8}\\
      6          &68.9    &85.7\\
    \hline
    \end{tabular}
    \label{tab:chunking number}
\end{table}

\textbf{The impacts of shared Transformers.}
Considering the computational efficiency, it is crucial to evaluate the configuration of Transformer blocks in LMF.
As illustrated in Tab.~\ref{tab:transformer setting}, we conduct several experiments with different settings.
Specifically, Model-1 indicates that we divide the feature map and utilize an average pooling to obtain local feature representations.
Model-2 and Model-3 mean that we employ one and two unshared Transformer layers to extract the local feature representation, respectively.
Model-4 and Model-5 means that we employ one and two shared Transformer layers to extract the local feature representation, respectively.
As can be observed, Model-5 achieves the best performance.
We believe that the improvements can be attributed to the fusion of multi-layer semantic features.
In addition, more shared Transformers enable the extraction of diverse information from multiple regions.
%
\begin{table}[hbtp]
      \centering
      \caption{Performance comparison with different settings in LMF.}
      \begin{tabular}{c|ccc|cc}
      \hline
               Model &Trans  &Shared   &Layer  &mAP  &Rank1\\
      \hline
       1&$\times$ &$\times$ &0      &68.2    &85.5\\
       2&\checkmark &$\times$ &1    &68.9    &85.7\\
       3&\checkmark &$\times$   &2   &67.6    &85.1\\
       4&\checkmark &\checkmark &1   &68.5    &85.5\\
       5&\checkmark &\checkmark &2   &\textbf{69.0}    &\textbf{85.8}\\
       \hline
      \end{tabular}
      \label{tab:transformer setting}
  \end{table}

\textbf{The impacts of different heads of GMA.}
To acquire more discriminative token representations, we introduce multiple heads in GMA.
To verify the impacts, we conduct several experiments, as presented in Tab.~\ref{tab: head of G2L}.
As can be observed, as the number of heads increases, the performance improves.
The results indicate that the multi-head attention mechanism can extract multiple semantic information and enhance the guided effect of global information.
\begin{table}[hbtp]
    \centering
    \caption{Performance comparison with different heads of GMA.}
    \begin{tabular}{c|cc}
    \hline
      Number  &mAP &Rank1\\
    \hline
      1          &68.5    &85.4\\
      4          &68.6    &85.4\\
      6          &68.8    &85.3\\
      12         &\textbf{69.0}    &\textbf{85.8}\\
    \hline
    \end{tabular}
    \label{tab: head of G2L}
\end{table}
\subsection{Qualitative Results}
\subsubsection{Visualization of Activation Maps}
To qualitatively analyze our model, we present the visual results of ViT~\citep{vit}, PCB$^{*}$ and our GLTrans on the MSMT17 dataset.
As illustrated in the second row of Fig.~\ref{fig:heatmap}, the scattered salient regions suggest that the ViT usually focuses on backgrounds and unrelated objects.
In contrast, the incorporation of part-based mechanisms can encourage the model to emphasize regions more pertinent to the target individual.
For instance, as depicted in Fig.~\ref{fig:heatmap}(e), (i), (k) and (l), the PCB$^{*}$ model discerns more semantically meaningful cues.
However, Fig.~\ref{fig:heatmap}(c), (f) and (h) demonstrate that PCB$^{*}$ relies on a few salient regions and miss useful cues due to the absence of global guidance.
Furthermore, our GLTrans demonstrates the effect of global guidance, allowing the model to focus on more discriminative and holistic information related to the person.
For example, as illustrated in Fig.~\ref{fig:heatmap}(f) and (i), our GLTrans not only identifies meaningful local regions but also mitigates the influence of extraneous information.
\subsubsection{Visualization of Retrieval Results}
\begin{figure}[htbp]
    \centering
    \includegraphics[width=1\linewidth]{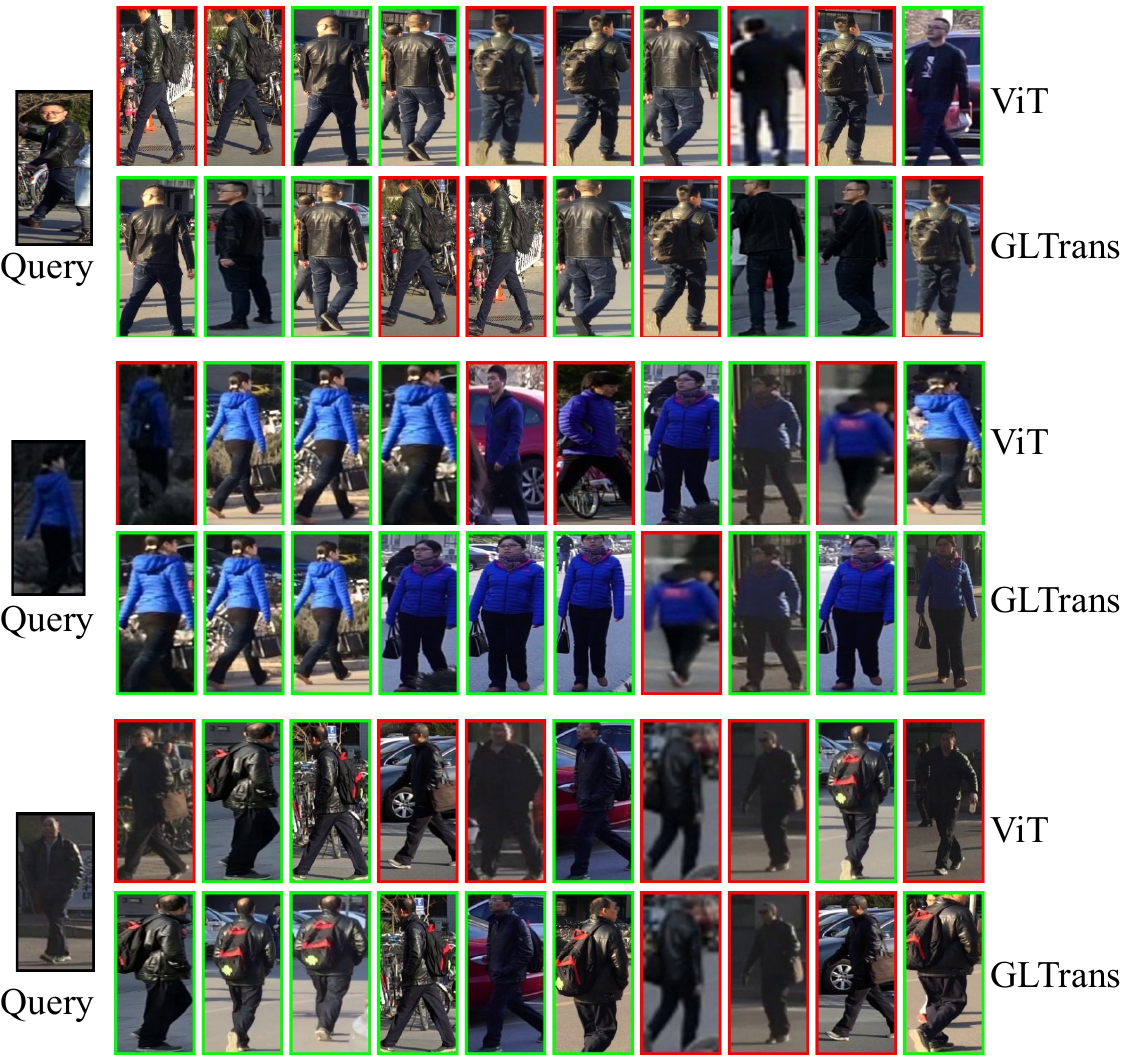}
    \caption{Retrieval results with three query samples on MSMT17. For each query, the first and second row are the ranking list produced by ViT and our GLTrans, respectively.
    The black, green and red boxes mean the query sample, true positive and false positive, respectively.}
    \label{fig:rankinglist}
    \vspace{-4mm}
\end{figure}

To further demonstrate the superiority of our proposed model, we present the retrieval results obtained by the ViT and our GLTrans.
As shown in Fig.~\ref{fig:rankinglist}, our GLTrans can get more hard positive samples and discard hard negative ones even if the global appearances of persons are similar.
For example, in the 3-\emph{th} row, the baseline model only concerns the person with a blue coat and black pants.
It misses the local cues (\emph{e.g.} gender and handbag).
As shown in the 4-\emph{th} row, our GLTrans obtains more robust and complementary representations, by refining both the global and local features.

Meanwhile, as shown in the 5-\emph{th} row, the baseline model can only retrieve the samples with the black coat and pants.
It misses local cues about the orange knapsack.
The main reason is that the baseline focuses on the global-view information and ignores the discriminative local features.
In contrast, our GLTrans achieves better results.
It considers both the global information from class tokens and the local details from patch tokens.
The same phenomenon can be found in other examples.
These visual results clearly illustrate that our GLTrans exploits both local and global information more effectively.
\section{Conclusion}
In this paper, we propose a novel learning framework named GLTrans for image-based object Re-ID.
To obtain complementary global representations, we propose a Global Aggregation Encoder (GAE) to aggregate multi-layer class tokens of ViT.
To mine discriminative local cues, we introduce a Local Multi-layer Fusion (LMF).
It contains three main modules, \emph{i.e.}, Patch Tokens Fusion (PTF), Global-guided Multi-head Attention (GMA) and Part-based Transformer Layers (PTL).
The PTF can adaptively enhance and fuse multi-layer patch tokens.
The GMA enhance the local patch tokens guided by the global-view.
The PTL adopts a local Transformer, encouraging more diverse and complementary local information.
The proposed framework takes the global and local advantages of vision Transformers for robust object Re-ID.
Experiments on four large-scale object Re-ID benchmarks demonstrate that our method achieves better performance than most state-of-the-art methods.
In the future, we will reduce the computation and improve the representation ability of diverse Transformers.

\bibliographystyle{model2-names}
\bibliography{refs}

\end{document}